\documentclass[conference]{IEEEtran}

\usepackage{amsmath}
\usepackage{graphicx}
\usepackage{booktabs}
\usepackage{multirow}
\usepackage[table]{xcolor}
\usepackage{tikz}
\usepackage{pgfplots, pgfplotstable}    
\usepackage{hyperref}
\usepackage{tablefootnote}

\newcommand{\Sref}[1]{\S\ref{#1}}
\newcommand{\Fref}[1]{Figure~\ref{#1}}
\newcommand{\Tref}[1]{Table~\ref{#1}}

\makeatletter
\def\blfootnote{\xdef\@thefnmark{}\@footnotetext}
\makeatother

\begin{document}

\title{Improving Multilingual ASR in the Wild\\Using Simple N-best Re-ranking}

\author{\IEEEauthorblockN{Brian Yan$^*$}
\IEEEauthorblockA{
\textit{Carnegie Mellon University}\\
byan@cmu.cs.edu \\
}
\and
\IEEEauthorblockN{Vineel Pratap}
\IEEEauthorblockA{
\textit{Meta FAIR}\\
vineelkpratap@meta.com}
\and
\IEEEauthorblockN{Shinji Watanabe}
\IEEEauthorblockA{
\textit{Carnegie Mellon University}\\
swatanab@andrew.cmu.edu \\
}
\and
\IEEEauthorblockN{Michael Auli}
\IEEEauthorblockA{
\textit{Meta FAIR}\\
michaelauli@meta.com
}
}

\maketitle

\begin{abstract}
Multilingual Automatic Speech Recognition (ASR) models are typically evaluated in a setting where the ground-truth language of the speech utterance is known, however, this is often not the case for most practical settings. 
Automatic Spoken Language Identification (SLID) models are not perfect and misclassifications have a substantial impact on the final ASR accuracy.
In this paper, we present a simple and effective N-best re-ranking approach to improve multilingual ASR accuracy for several prominent acoustic models by employing external features such as language models and text-based language identification models. 
Our results on FLEURS using the MMS and Whisper models show spoken language identification accuracy improvements of 8.7\% and 6.1\%, respectively and word error rates which are 3.3\% and 2.0\% lower on these benchmarks.
The code is available at: \url{https://github.com/facebookresearch/fairseq/tree/main/examples/mms/lid_rerank}.
\end{abstract}

\IEEEpeerreviewmaketitle

\section{Introduction}
\label{sec:intro}

Recent work in multilingual automatic speech recognition (ASR) has drastically increased language coverage to support hundreds and even thousands of languages. 
This includes approaches based on labeled training data such as Whisper \cite{radford2022whisper}, USM \cite{zhang2023usm}, Seamless~\cite{communication2023seamless} and MMS \cite{pratap2024mms} as well as zero-shot work~\cite{li2022asr2k,zhao2024mmszero}.
The typical evaluation setting of these models assumes that the ground-truth language is known in advance to the model and we refer to this as the \textit{lab} evaluation setting.
However, in most practical settings, the language is not known and needs to be predicted using spoken language identification (SLID) - a setting we refer to as the \textit{wild}.
\blfootnote{$^*$Work done during an internship at Meta FAIR.}

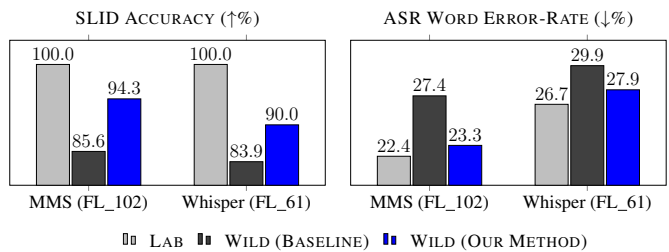
\begin{figure}[t]
\centering
\resizebox {\linewidth} {!} {
\begin{tikzpicture}

  \begin{scope}[xshift=0cm]
    \begin{axis}[
      ybar,
      width=0.55\textwidth,
      height=0.3\textwidth,
      xmin=.5,
      xmax=2.5,
      ymin=80,
      ymax=104,
      ytick=\empty,
      xtick={1,2},
      xticklabels={\Large MMS (FL\_102), \Large Whisper (FL\_61)},
      bar width=25,
      title=\Large \textsc{SLID Accuracy ($\uparrow$\%)},
      legend style={at={(1.1,-0.3)}, anchor=north, draw=none, fill=none, style={column sep=0.3cm}},
      legend columns=3,
      nodes near coords,
      every node near coord/.append style={font=\Large},
    ]
    \addlegendentry{\Large \textsc{Lab}}
    \addlegendentry{\Large \textsc{Wild (Baseline)}}
    \addlegendentry{\Large \textsc{Wild (Our Method)}}
    \addplot[draw=black,fill=lightgray] coordinates {
      (1, 100)
      (2, 100)
    };
    \addplot[draw=black,fill=darkgray] coordinates {
      (1, 85.6)
      (2, 83.9)
    };
    \addplot[draw=black,fill=blue] coordinates {
      (1, 94.3)
      (2, 90.0)
    };
    \pgfkeys{/pgf/number format/.cd,fixed,fixed zerofill,precision=1}
    \end{axis}
  \end{scope}
  
  \begin{scope}[xshift=0.5\textwidth]
    \begin{axis}[
      ybar,
      width=0.55\textwidth,
      height=0.3\textwidth,
      xmin=.5,
      xmax=2.5,
      ymin=20,
      ymax=32,
      ytick=\empty,
      xtick={1,2},
      xticklabels={\Large MMS (FL\_102), \Large Whisper (FL\_61)},
      bar width=25,
      title=\Large \textsc{ASR Word Error-Rate ($\downarrow$\%)},
      nodes near coords,
      every node near coord/.append style={font=\Large},
    ]
    \addplot[draw=black,fill=lightgray] coordinates {
      (1, 22.4)
      (2, 26.7)
    };
    \addplot[draw=black,fill=darkgray] coordinates {
      (1, 27.4)
      (2, 29.9)
    };
    \addplot[draw=black,fill=blue] coordinates {
      (1, 23.3)
      (2, 27.9)
    };
    \end{axis}
  \end{scope}
\end{tikzpicture}
}
\vspace{-10mm}
\caption{Multilingual ASR evaluation often assumes perfect spoken language identification (Lab) leading to much lower word error rates compared to real world spoken language identification (Wild - Baseline).
Our method alleviates this lab-to-wild degradation (Wild - Our Method).
}
\vspace{-4mm}
\label{fig:lab_v_wild}
\end{figure}

SLID models~\cite{radford2022whisper,pratap2024mms} perform well on high-resource languages such as English but less so on low-resource languages, e.g., for Whisper, there are 13 out of 100 supported languages which have a SLID accuracy of below 50\% on FLEURS~\cite{conneau2023fleurs}.
The errors of SLID bias dowstream ASR models towards the incorrect language and may result in unusable outputs.
This typically results in a higher average word error rate for ASR but it is important to note that this increase is driven by high error rates on a subset of samples for which the output will be essentially unusable.
Figure~\ref{fig:lab_v_wild} illustrates the impact of imperfect SLID on ASR performance and how our method can recover much of that lost performance.

\begin{figure}[t]
\centering
\includegraphics[width=0.6\linewidth]{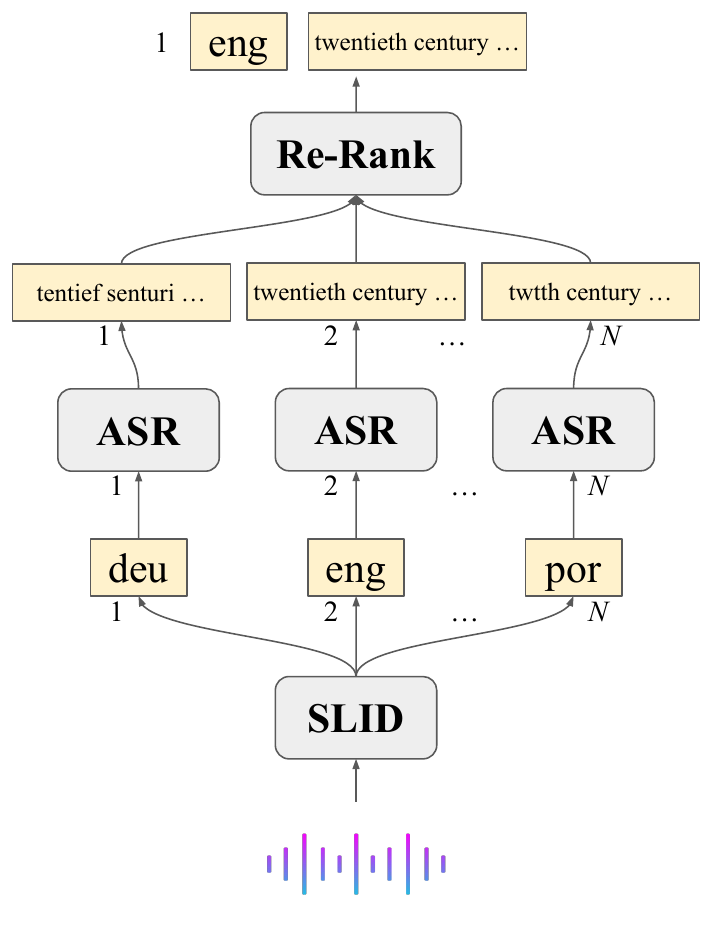}
\vspace{-4mm}
\caption{Illustration of our multilingual $N$-best re-ranking approach.
}
\label{fig:summary}
\vspace{-4mm}
\end{figure}

In this paper, we propose a simple and effective $N$-best re-ranking approach to alleviate the error propagation from SLID to ASR (\Sref{sec:proposed}).
The key idea is to defer the final SLID decision until \emph{after} ASR has been performed for the top $N$ SLID predictions. 
This enables the use of features operating over the resulting transcriptions to better determine the correct language. 
The set of features includes language models, written language identification models, as well as simple acoustic models to choose the final output (Figure~\ref{fig:summary} illustrates our method).

The results show that this method can substantially improve accuracy on both FLEURS~\cite{conneau2023fleurs} and ML-SUPERB~\cite{shi2023ml} for MMS and Whisper, as well as Seamless.
Moreover, the method often approaches the ASR performance which could have been achieved with  the correct SLID prediction.
A downside of the method is the higher computational requirements, however, we find that even for small N-best list sizes of $N=2$, the majority of the accuracy improvements can be realized.

\section{Proposed Method}
\label{sec:proposed}

We adapt the well-known $N$-best re-ranking framework typically used in monolingual ASR to multilingual ASR (\Sref{sec:method_multilingual}).
We then describe the set of quality estimation systems used to score and re-rank the multilingual $N$-best list (\Sref{sec:method_features}).

\subsection{Multilingual ASR N-best Lists}
\label{sec:method_multilingual}
$N$-best list re-ranking for a single language operates by using the ASR model to generate $N$-best candidate transcriptions and ranking them by their likelihood, for instance via beam search or sampling \cite{och-2003-minimum, chan2016LAS, salazar2020masked, li2022recent, prabhavalkar2024survey}.
Finally, the candidates are then re-scored according to additional features and a new candidate is chosen.
Crucially, the $N$ candidate transcriptions are all in the same language.

The \textbf{multilingual $N$-best re-ranking} we consider operates differently:
first, we perform spoken language identification to obtain the $N$ highest scoring languages predicted by the SLID model for a given utterance.
Next, we obtain the single highest scoring transcription for each of the $N$ languages from the multilingual ASR system by conditioning the ASR model on each language. 
Finally, we place the resulting transcription for each language in the $N$-best list for re-ranking.

\subsection{Features for Re-ranking}
\label{sec:method_features}

We utilize the following six \textbf{features for re-ranking}, each of which estimates a subset of the overall joint likelihood of a speech-language-transcript triplet, $P(S,L,T)$:
\begin{itemize}
    \item $P(T)$ : (1) Language Model (LM), (2) Text Length (Len)\footnote{Not an actual likelihood estimator but can be used as a (weak) proxy; for instance, 0-length or max-length may indicate low likelihood.}
    \item $P(L,T)$ : (3) Written LID Model (WLID)
    \item $P(S,L)$ : (4) Spoken LID Model (SLID)
    \item $P(S,T)$ : (5) ASR, (6) u-roman ASR (UASR)
\end{itemize}
where the SLID and ASR systems are the exact same systems which produced the $N$-best list, i.e., we do not use Whisper to re-rank MMS or vice-versa.

These features (aside from SLID) judge the quality of the transcripts $T$ in different ways:
the LM judges fluency of $T$ in the respective language, the WLID model operates similarly but is more focused on individual words. Both can identify ASR outputs consisting of many non-word tokens such as when SLID is incorrect.
u-roman ASR~\cite{zhao2024mmszero} is an acoustic CTC model trained on romanized text \cite{hermjakob2018out} which standardizes the text of any language to the standard Latin alphabet using a set of heuristics. 
The u-roman acoustic model penalizes candidates which are not faithful to the audio and we found this to be useful for ASR models known to hallucinate such as Whisper.

\section{Experimental Setup}
\label{sec:setup}

\subsection{Spoken Lang ID and Automatic Speech Recognition Models}

\Tref{tab:models} shows the three multilingual ASR setups we chose for our study which are based on recent prominent open-sourced models.
For Whisper \cite{radford2022whisper}, a single auto-regressive neural network first predicts SLID via a language token (e.g. [eng] or [spa]) which conditions the generation of subsequent ASR tokens.
Joint prediction of SLID and ASR is a common approach in multilingual ASR~\cite{watanabe2017joint, toshniwal2018, hou2020, chen2023improving}.

For MMS and Seamless, separate classification-based SLID models first predict SLID. 
Then MMS \cite{pratap2024mms} takes the language prediction by selecting language-specific CTC adapters, while Seamless ~\cite{communication2023seamless} auto-regressively conditions ASR on a language token similar to Whisper.

We use greedy decoding for all ASR models. We do not use an external LM to decode the CTC-based MMS model.

\begin{table}[t]
  \centering
    \caption{We test our method on three separate model setups, each consisting of Spoken Language Identification (SLID) and Automatic Speech Recognition (ASR) components.
    }
    \vspace{-4mm}
    \resizebox {\linewidth} {!} {
\setlength{\tabcolsep}{3.5pt}
\begin{tabular}{lcll}
\toprule
\textsc{Name} & \textsc{Langs} & \textsc{SLID Component} & \textsc{ASR Component} \\
\midrule
MMS \cite{pratap2024mms} & 1162 & MMS-1B\tablefootnote{\url{https://huggingface.co/facebook/mms-lid-4017}; We use a slightly different version of this publicly availabel model - ours removes FLEURS training data because in-domain performance is not representative of that in the wild.}  & MMS-1B\tablefootnote{\url{https://huggingface.co/facebook/mms-1b-all}}\\
Whisper \cite{radford2022whisper} & 100 & Whisper v2 & Whisper v2\tablefootnote{\url{https://huggingface.co/openai/whisper-large-v2}} \\
Seamless \cite{communication2023seamless} & 96 & ECAPA-TDNN\tablefootnote{\url{https://huggingface.co/speechbrain/lang-id-voxlingua107-ecapa}} \cite{speechbrain} & SeamlessM4T v2\tablefootnote{\url{https://huggingface.co/facebook/seamless-m4t-v2-large}}  \cite{communication2023seamless} \\
\bottomrule
\end{tabular}
}

    \label{tab:models}
    \vspace{-4mm}
\end{table}

\subsection{Re-ranking Feature Models}
We use 6 auxiliary re-ranking models to produce the 6 features for re-ranking introduced in \Sref{sec:method_features}:
\begin{itemize}
    \item $P(T)$ : (1) For LM we use MaLA-500 \cite{lin2024mala} which supports 534 languages and is based on Llama-2 7B. 
(2) For Len we use character count, including spaces.
    \item $P(L,T)$ : (3) For WLID we use NLLB \cite{costa2022no} which can classify text into 200 languages.
    \item $P(S,L)$ : (4) For SLID we use the same model which produced the $N$-best languages in the particular setup.
    \item $P(S,T)$ : (5) For ASR we use the same model which produced the $N$-best list in the particular setup. (6) Finally, for UASR we use MMS\_Zero-Shot \cite{zhao2024scaling} which supports force alignment with romanized text.
\end{itemize}
All of these models produce log likelihood scores with the exception of Text Length (Len).

\subsection{Hyper-parameter Tuning}
The re-ranking is determined by the combined feature score:
\begin{equation}
    \text{score} = \sum_{i=1}^{F} w_i f_i
\end{equation}
where $f_i$ is the feature value of one of the $F=6$ aforementioned models and $w_i$ is the feature weight coefficient.
The weights are tuned via random search over 10k iterations in which each coefficient is randomly set within a defined range.
This is achieved by computing the re-ranking WER on a development set for all 10k coefficient sets and then selecting the one which minimizes WER.
We use the official dev sets of FLEURS and ML-SUPERB, tuning separately for each model and dataset.
For LM, ASR, and UASR we use a range of 0 to 10.
For SLID and WLID we use a range of 0 to 100.
For Text Length we use a range of -5 to 5.

\subsection{Datasets and Evaluation}
We evaluate on FLEURS (FL;~\cite{conneau2023fleurs}) and ML-SUPERB (MS;~\cite{shi2023ml}) using the languages supported by each model.
For instance, MMS is evaluated on FL\_102, the full benchmark of 102 languages, while Whisper is evaluated on FL\_61, the supported subset.
Therefore, MMS results are not directly comparable to Whisper. 
We made this choice to evaluate our approach for each model as broadly as possible.

We report both SLID and ASR performance, where the former is measured in terms of accuracy and the latter is measured in terms of  word error rate (WER) for all languages except character-based languages for which character error rate is used.\footnote{Character langs: adx, bod, cmn, dzo, jpn, khg, khm, lao, mya, tha, yue.}
We compare our method to the baseline 1-best SLID+ASR performance as well as the topline (oracle) re-ranking performance. 
The oracle is based on selecting the candidate with the correct language if it exists in the $N$-best list. 
If the correct language is not in the $N$-best list, then the candidate with the highest SLID score is selected.

\section{Results and Analysis}
\label{sec:results}

We first report the accuracy of $N$-best re-ranking on Whisper, MMS, and Seamless both for SLID and ASR (\Sref{sec:results_main}), analyze potential regressions (\Sref{sec:results_regressions}), and improvements on tail langauges (\Sref{sec:results_taillang}).
Next, we ablate the relative importance of each re-ranking feature (\Sref{sec:results_importance}).
Finally, we show that our method is still effective with small values of $N$ (\Sref{sec:results_n}) and outperforms standard monolingual $N$-best re-ranking (\Sref{sec:standard_nbest}).

\subsection{Main Results}
\label{sec:results_main}

\begin{table}[t]
  \centering
    \caption{SLID Accuracy ($\uparrow$\%): Re-rank 10-best (our method) vs. baseline (1-best) and oracle (10-best). 
    }
    \vspace{-4mm}
    \resizebox {\linewidth} {!} {
\setlength{\tabcolsep}{3.5pt}
\begin{tabular}{l|cc|cc|cc|c}
\toprule
& \multicolumn{2}{c|}{\textsc{MMS}} & \multicolumn{2}{c|}{\textsc{Whisper}} & \multicolumn{2}{c|}{\textsc{Seamless}} & Avg \\
\midrule
& FL\_102 & MS\_143 & FL\_61 & MS\_76 & FL\_73 & MS\_88 & \\
\midrule
Baseline & 85.6 & 80.3 & 83.9 & 76.7 & 81.4 & 69.6 & 79.6\\
Re-rank & 94.3 & 85.0 & 90.0 & 79.5 & 83.1 & 75.8 & 84.6\\
Oracle & 96.7 & 93.0 & 98.6 & 92.7 & 86.8 & 83.6 & 91.9\\
\bottomrule
\end{tabular}
}
    \label{tab:main_lid}
    \vspace{-4mm}
\end{table}

\begin{table}[t]
  \centering
    \caption{ASR Word Error Rate ($\downarrow$\%): Re-rank 10-best (our method) vs. baseline (1-best) and oracle (10-best).}
    \vspace{-4mm}
    \resizebox {\linewidth} {!} {
\setlength{\tabcolsep}{3.5pt}
\begin{tabular}{l|cc|cc|cc|c}
\toprule
 & \multicolumn{2}{c|}{\textsc{MMS}} & \multicolumn{2}{c|}{\textsc{Whisper}} & \multicolumn{2}{c|}{\textsc{Seamless}} & Avg \\
\midrule
System & FL\_102 & MS\_143 & FL\_61 & MS\_76 & FL\_73 & MS\_88 & \\
\midrule
Baseline & 26.6 & 29.7 & 29.9 & 33.8 & 28.5 & 32.8 & 30.2 \\
Re-rank & 23.3 & 26.9 & 27.9 & 30.9 & 26.2 & 28.9 & 27.4\\
Oracle & 23.0 & 26.4 & 27.7 & 30.6 & 25.2 & 26.6 & 26.6 \\
\bottomrule
\end{tabular}
}
    \label{tab:main_asr}
    \vspace{-4mm}
\end{table}

\Tref{tab:main_lid} shows that the method \textbf{increases SLID accuracy on average by 5\% absolute} on FLEURS and ML-SUPERB and three model setups.
The approach captures about 41\% of the oracle performance.

\Tref{tab:main_asr} shows the corresponding ASR performance. 
10-best re-ranking \textbf{reduces the WER by nearly 3\% absolute}, or about 9\% relative, and captures about 79\% of the oracle which has access to the real language ID.

Generally, performance is better on FLEURS than ML-SUPERB which we attribute to the latter being a harder benchmark.
The greater improvements in ASR than in SLID are due only in part to our choice to minimize ASR WER during coefficient tuning.
Another factor is that in some cases a model can produce better outputs (lower WER) for an incorrect language than for the correct one, but this is rare (more often on languages with high WER in the lab setting).

\subsection{Drilldown into Main Results}
\label{sec:results_regressions}

\begin{table}[t]
  \centering
    \caption{Performance Breakdown into originally correct SLID vs. originally incorrect SLID for MMS FL\_102.
    There is only a very small regression on the correct subset and large improvements on the incorret subset.
    }
    \vspace{-4mm}
\setlength{\tabcolsep}{3.5pt}
\begin{tabular}{l|cc|cc}
\toprule
& \multicolumn{2}{c|}{\textsc{Orig. Correct SLID}} & \multicolumn{2}{c}{\textsc{Orig. Incorrect SLID}} \\
& \multicolumn{2}{c|}{\textsc{(85.6\% of total)}} & \multicolumn{2}{c}{\textsc{(14.4\% of total)}} \\
\midrule
System & Acc ($\uparrow$\%) & WER ($\downarrow$\%) & Acc ($\uparrow$\%) & WER ($\downarrow$\%) \\
\midrule
Baseline & 100.0 & 20.8 & 0.0 & 69.1 \\
Re-rank & 99.0 & 20.9 & 66.7 & 40.4 \\
Oracle & 100.0 & 20.8 & 76.9 & 39.3 \\
\bottomrule
\end{tabular}
    \label{tab:subset}
    \vspace{-4mm}
\end{table}

The results so far showed an overall improvement but is there any regression on the samples which were originally correctly classified by SLID?

To better understand this issue, we divide SLID and ASR performance into samples classified correctly and incorrectly by the original SLID.
\Tref{tab:subset} shows that the correct subset has slight SLID accuracy regression of 1\% and a 0.1\% WER degradation for ASR, while the SLID accuracy of the error subset increased from 0\% to nearly 67\% with a WER reduction of nearly 29\% absolute. 
It is important to note that the correct subset is much larger (85.6\% vs. 14.4\%) and regressions there have a larger impact.
Nonetheless, the regression is still very small and is accompanied by substantial improvements on the error subset.
Moreover, the oracle shows that the error subset is hard: the oracle accuracy of 39.3\% WER is nearly double that of the correct subset (20.8\%).

\subsection{Effect on Tail Languages}
\label{sec:results_taillang}

\begin{table}[t]
  \centering
    \caption{Tail Performance: Re-ranking impact on the ten lowest performing languages (by SLID Acc) for MMS FL\_102.}
    \vspace{-4mm}
\begin{tabular}{l|cc|cc}
\toprule
 & \multicolumn{2}{c|}{\textsc{SLID Acc ($\uparrow$\%)}} & \multicolumn{2}{c}{\textsc{ASR WER ($\downarrow$\%)}}  \\
\midrule
Language & Baseline & Re-rank & Baseline & Re-rank \\
\midrule
Xhosa & 0.0 & 76.1 & 56.4 & 36.2 \\
Kabuverdianu & 1.2 & 40.4 & 52.4 & 34.4 \\
Irish &	5.3 & 96.0 & 96.9 & 62.4 \\
Occitan & 13.6 & 96.2 & 48.8 & 34.5 \\
Odia & 18.4 & 93.7 & 87.1 & 39.6 \\
Kyrgyz & 23.6 & 93.2 & 32.8 & 19.0 \\
Central Kurdish & 28.0 & 86.1 & 74.3 & 44.8 \\
Northern Sotho & 29.6 & 99.4 & 75.3 & 27.3 \\
Igbo & 30.3 & 63.5 & 72.7 & 52.0 \\
Umbundu & 30.7 & 86.4 & 77.3 & 42.8 \\
\midrule
Average & 18.1 & 83.1 & 67.4 & 39.3 \\
\bottomrule
\end{tabular}
    \label{tab:tail}
    \vspace{-4mm}
\end{table}

The baseline SLID accuracy is not uniform across languages - does the method alleviate the problems at the tail?

\Tref{tab:tail} shows the SLID and ASR performance for the ten FLEURS languages with the lowest baseline SLID accuracy for MMS as well as the improvements with our method.
Our method increases the average SLID accuracy from 18.1\% to 83.1\% and decreases average WER from 67.4\% to 39.3\%.
See for instance the Odia language - SLID accuracy increases by 75\% absolute while WER decreases by 48\% absolute.

\subsection{Feature Ablation}
\label{sec:results_importance}

To better understand the relative importance of each of the six features we use for re-ranking, we conduct the following experiment: 
starting from an empty set we add the feature which results in the best performance after tuning coefficients to minimize WER, repeating until all features have been added.
The rankings for MMS FL\_102 and Whisper FL\_61 are reported in \Tref{tab:ablation}.

\begin{table}[t]
  \centering
    \caption{Feature Importance Rankings: MMS FL\_102 and Whisper FL\_61.}
    \vspace{-4mm}
\begin{tabular}{clc|clc}
\toprule
\multicolumn{3}{c|}{\textsc{MMS (FL\_102)}} & \multicolumn{3}{c}{\textsc{Whisper (FL\_61)}} \\
\midrule
Rank & Signal & WER ($\downarrow$) & Rank & Signal & WER ($\downarrow$) \\
\midrule
1 & SLID & 27.0 & 1 & SLID & 29.9 \\
2 & +WLID & 25.1 & 2 & +UASR & 29.2 \\
3 & +ASR & 24.5 & 3 & +ASR & 28.8 \\
4 & +LM & 24.1 & 4 & +WLID & 28.0 \\
5 & +Len & 23.7 & 5 & +Len & 28.0 \\
6 & +UASR & 23.7 & 6 & +LM & 27.9 \\
\bottomrule
\end{tabular}
    \label{tab:ablation}
    \vspace{-4mm}
\end{table}

The most important feature for both MMS and Whisper is SLID which is in part because it is the model which originally produces the list of the $N$-best languages.
The remaining feature rankings show some variation across models which we believe is due to the different nature of the models and their different error patterns.
For instance, the uroman acoustic model (UASR) is the second most useful feature for Whisper but the least useful for MMS. 
This is likely because Whisper tends to generate unrelated transcriptions and UASR is effective in determining this since UASR is a CTC model which is force-aligned with the Whisper transcription. 
On the other hand, MMS is already a CTC model and this capability is therefore redundant.

Written LID (WLID) is the second most important feature for MMS because the CTC model is more likely to output non-words given the incorrect language tag and WLID detects this effectively.
The acoustic model score (ASR) is useful for both models since it is measure of confidence in the transcriptions.
Finally, an external language model is most redundant for Whisper, since it has a strong built-in language model already.

\subsection{$N$-best List Size Ablation}
\label{sec:results_n}

A drawback of the method is an increase in computational cost proportional to the size of the $N$-best list. 
To better understand the impact of $N$, we measure performance for different values (where $N=1$ corresponds to the baseline).

\begin{figure}[t]
\centering
\resizebox {\linewidth} {!} {
\begin{tikzpicture}

  \begin{scope}[xshift=0cm]
    \begin{axis}[
    title=\Large \textsc{SLID Accuracy ($\uparrow$\%)},
    xlabel=\Large $N$-best,
    xmin=0.5,
    xmax=10.5,
    ymin=83,
    ymax=96,
    xtick={1,2,3,4,5,6,7,8,9,10},
    grid=major,
    grid style={dashed},
    legend cell align={left},
    legend style={at={(1.1,-0.25)}, anchor=north, draw=none, fill=none, style={column sep=0.3cm}},
    legend columns=2,
    x tick label style={font=\large},
    y tick label style={font=\large},
  ]
  
  \addplot[blue,ultra thick,mark=square*] coordinates {
    (1,85.6)
    (2,89.6)
    (3,91.2)
    (4,92.4)
    (5,93.1)
    (6,93.7)
    (7,94.0)
    (8,94.1)
    (9,94.2)
    (10,94.3)
  };
  
  \addplot[black,ultra thick,mark=otimes] coordinates {
    (1,83.9)
    (2,87.2)
    (3,88.0)
    (4,88.6)
    (5,89.1)
    (6,89.5)
    (7,89.7)
    (8,89.8)
    (9,90.0)
    (10,90.0)
  };

    \addlegendentry{\Large \textsc{Re-rank (MMS FL\_102)}}
    \addlegendentry{\Large \textsc{Re-rank (Whisper FL\_61)}}
  
  \end{axis}
  \end{scope}
  
  \begin{scope}[xshift=0.45\textwidth]
    \begin{axis}[
    title=\Large \textsc{ASR Word Error-Rate ($\downarrow$\%)},
    xlabel=\Large $N$-best,
    xmin=0.5,
    xmax=10.5,
    ymin=22,
    ymax=31,
    xtick={1,2,3,4,5,6,7,8,9,10},
    grid=major,
    grid style={dashed},
    x tick label style={font=\large},
    y tick label style={font=\large},    
  ]
  
  \addplot[blue,ultra thick,mark=square*] coordinates {
    (1,26.6)
    (2,24.6)
    (3,24.0)
    (4,23.7)
    (5,23.6)
    (6,23.4)
    (7,23.4)
    (8,23.3)
    (9,23.3)
    (10,23.3)
  };

  \addplot[black,ultra thick,mark=otimes] coordinates {
    (1,29.9)
    (2,28.4)
    (3,28.1)
    (4,28.0)
    (5,27.9)
    (6,27.9)
    (7,27.9)
    (8,27.9)
    (9,27.9)
    (10,27.9)
  };

  \end{axis}
  \end{scope}
\end{tikzpicture}
}
\vspace{-10mm}
\caption{Effect of different sized $N$-best lists for SLID (left) and ASR (right). 
}
\label{fig:slid_curve}
\end{figure}
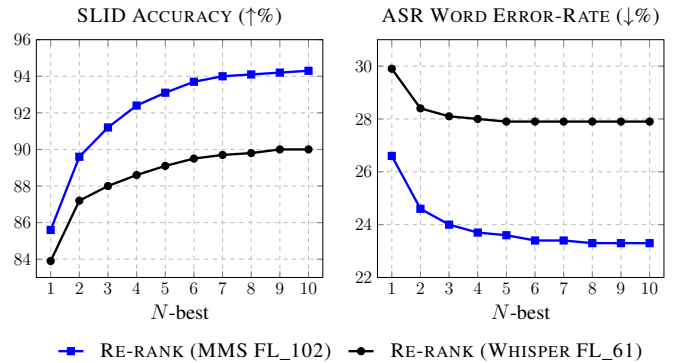

\Fref{fig:slid_curve} shows that the larger values of $N$ result in better accuracy but the largest marginal improvement is achieved with just $N=2$ which captures 46\% and 61\% of the SLID and ASR improvements of $N=10$ for MMS and 55\% and 75\% for Whisper, respectively.
This suggests that a lightweight version of the proposed method may be preferable if computational cost is a limiting factor.

\subsection{Comparison to Standard Monolingual N-best Lists}
\label{sec:standard_nbest}

\begin{table}[t]
  \centering
    \caption{Multilingual vs. Monolingual N-best Lists: Whisper FL\_61}
    \vspace{-4mm}
\begin{tabular}{l|cc}
\midrule
System & Acc ($\uparrow$\%) & WER ($\downarrow$\%)  \\
\midrule
Baseline (1-best) & 83.9 & 29.9 \\
Monolingual Re-rank (10-best) & 83.9 & 28.7 \\
Multilingual Re-rank (10-best) & 90.0 & 27.9 \\
\bottomrule
\end{tabular}
    \label{tab:standard_nbest}
    \vspace{-4mm}
\end{table}

Finally, we compare multilingual re-ranking to standard monolingual re-ranking which is based on an $N$-best list obtained via beam search for the most likely LID prediction of an utterance.
For this comparison we use the same set of re-ranking features for each type of $N$-best list.

\Tref{tab:standard_nbest} shows that monolingual re-ranking outperforms the 1-best baseline but our approach further improves ASR by 0.8\% absolute WER while also improving SLID accuracy by 6.1\% absolute. 
This shows that a large part of the improvement is due to identifying the correct language and not just due to switching to an n-best re-ranking setup.
Moreover, our method may also benefit from considering multiple ASR candidates per language and future work may investigate this.

\section{Related Work}

The general idea of determining language ID after examining ASR outputs is not new. Prior works have touched on this idea as part of broader explorations aimed at improving the LAS architecture \cite{toshniwal2018, li2018multi}.
Other works focused on code-switched ASR in particular have also explored various forms of SLID and ASR conditioning \cite{yan2022joint, yan2023towards, hussein2024}.

Other works are more directly similar to our core idea of considering a set of candidate languages for each utterance.
Our work builds upon these prior studies by offering a relatively simpler yet highly effective approach along with bench-marking on vastly more languages (thanks to modern open-source foundation models and data).
\cite{metze2000confidence} uses ASR confidence to distinguish between only three possible languages, an early precursor to our work.
\cite{wang2019signal} uses a set of features like we do but they propose either lattice regression or neural network based methods for combining them, both of which are more complex than our $N$-best re-ranking method.
Finally, \cite{chandak2020streaming} eschews the use of any features and instead learn latent representations of ASR candidates via a bespoke architecture; their method does not improve other existing multilingual ASR systems. 

\section{Conclusion}

This paper presents a simple $N$-best re-ranking approach to improve multilingual ASR using a small set of external features.
Empirical results show spoken language identification accuracy improvements with the Whisper and MMS models of 8.7\% and 6.1\%, respectively and word error rate reductions of 3.3\% and 2.0\% on the FLEURS benchmark.
Moreover, we show that our method substantially benefits many languages which previously had very low SLID and ASR performance.



\bibliographystyle{IEEEbib}
\bibliography{refs}

\end{document}